\documentclass[runningheads]{llncs}
\usepackage{amsmath,amsfonts}
\usepackage{algorithmic}
\usepackage{graphicx}
\usepackage{epstopdf}
\usepackage{epsfig}
\usepackage{array}
\usepackage[caption=false,font=normalsize,labelfont=sf,textfont=sf]{subfig}
\usepackage{textcomp}
\usepackage{stfloats}
\usepackage{url}
\usepackage{verbatim}
\usepackage{amsmath}
\usepackage[T1]{fontenc}
\usepackage{bm}
\usepackage[caption=false]{subfig}
\usepackage[colorlinks,bookmarksopen,bookmarksnumbered,citecolor=green, linkcolor=red, urlcolor=blue]{hyperref}
\usepackage{graphicx}
\usepackage{booktabs} 
\usepackage{caption}
\def\BibTeX{{\rm B\kern-.05em{\sc i\kern-.025em b}\kern-.08em
		T\kern-.1667em\lower.7ex\hbox{E}\kern-.125emX}}
\usepackage{balance}
\begin{document}
\title{Second-order Anisotropic Gaussian Directional Derivative Filters for Blob Detection}
%
%
\author{Jie Ren\inst{1} \and
Wenya Yu\inst{1} \and
Jiapan Guo\inst{2} \and
Weichuan Zhang\inst{3} \and
Changming Sun\inst{4}
}
\authorrunning{Jie et al.}
%
\institute{School of Electronics and Information, Xi'an Polytechnic University, Xi'an, China
\email{renjie@xpu.edu.cn; 210411053@stu.xpu.edu.cn}\\
 \and
Department of Radiation Oncology, University Medical Center Groningen, University of Groningen, Hanzeplein 1, 9713GZ Groningen, Netherlands \\
\email{j.guo@rug.nl}
 \and
Institute for Integrated and Intelligent Systems, Griffith University, QLD, Australia\\
\email{weichuan.zhang@griffith.edu.au; zwc2003@163.com}
 \and
CSIRO Data61, PO Box 76, Epping, NSW 1710, Australia\\
\email{Changming.Sun@data61.csiro.au}}
\maketitle              
\begin{abstract}
Interest point (corner/blob) detection methods have received increasing attention and are widely used in computer vision tasks such as image retrieval and 3D reconstruction. In this work, second-order anisotropic Gaussian directional derivative filters with multiple scales are used to smooth the input image and a novel blob detection method is proposed. Extensive experiments demonstrate the superiority of our proposed method over state-of-the-art benchmarks in terms of localization of interest point detection, shape description of detected interest points, and image matching.

\end{abstract}
\section{Introduction}
Interest points are key image features which have been widely used for visual tasks such as image matching, image retrieval, and object tracking~\cite{li2019multi,JING2022259,jing2021novel}. Currently, scholars~\cite{9866553,ren2020contour,wang2018survey,zhang2013image,lowe2004distinctive,zhang2015contour,zhang2019discrete} classify corners and blobs as interest points, and various interest point detection methods~\cite{lindeberg1993detecting,ono2018lff} have been presented in the literature. The existing interest point detection methods intend to extract corners or blobs using first-order image intensity variations~\cite{wang2020corner,shui2013corner,zhang2014corner,10026417}, second-order image intensity variations~\cite{9234393,lowe2004distinctive,8883063}, or machine learning based techniques~\cite{ono2018lff,tian2020d2d,jing2023ecfrnet}.

In this work, the second-order anisotropic Gaussian directional derivative (SOAGDD) filters with multiple scales are utilized for obtaining second-order anisotropic Gaussian directional derivatives with multiple scales from input images. Then a novel interest point detection method is proposed which not only has the capability to localize interest points but also has the capability to describe the shape of interest points. Extensive experiments demonstrate the superiority of our proposed method over state-of-the-art benchmarks based on the localization of interest point detection, shape description of detected interest points, and image matching.

\section{Related Work}
Currently, second-order horizontal and vertical directional isotropic Gaussian derivatives with multiple scales have been widely used for blob detection. In~\cite{beaudet1978rotational}, input image $I(x,y)$ was smoothed by an isotropic Gaussian filter $g_{\sigma}(x,y)$ with a scale factor $\sigma$ ($\sigma$$>$$0$)
\begin{equation}\begin{aligned}
\label{eq1}
\hbar(x,y) = I(x,y)\otimes g_{\sigma}(x,y),
\end{aligned}\end{equation}
where $(x,y)$ represents a pixel coordinate in the image. Then the second-order horizontal, vertical, and cross directional derivatives (${\hbar_{xx}(x,y)}$, ${\hbar_{xy}(x,y)}$, and ${\hbar_{yy}(x,y)}$) of ${\hbar(x,y)}$ are derived to construct an Hessian matrix
\begin{equation}
	H = \left[ {\begin{array}{*{20}{c}}
			{{\hbar_{xx}}}&{{\hbar_{xy}}} \\
			{{\hbar_{xy}}}&{{\hbar_{yy}}}
	\end{array}} \right].
\end{equation}
Interest points are defined as the local extremum (either a maximum or a minimum) of the determinant of the Hessian
\begin{equation}
	C = \left| {{H}(\hbar )} \right| = {\hbar_{xx}}{\hbar_{yy}} - {({\hbar_{xy}})^2}.
\end{equation}
In~\cite{lowe2004distinctive},  a scale invariant feature transform (SIFT) method was proposed which utilizes a difference of Gaussian (DoG) filter for obtaining second-order derivatives with multiple scales from an input image
\begin{equation}
	\begin{aligned}	
		D(x,y,\sigma)=(G(x,y,k*\sigma)-G(x,y,\sigma)){\otimes}I(x,y),
	\end{aligned}
\end{equation}
where ${k}$ is a constant. The candidate interest points are detected by seeking for local maxima in a DoG pyramid. Meanwhile, the eigenvalues (${\lambda _{\max }}$, ${\lambda _{\min }}$, with ${\lambda _{\max }}$ > ${\lambda _{\min }}$) of the Hessian matrix (with the second-order derivative extracted by DoG) are applied to suppress edge responses
\begin{equation}
	\left[ {\begin{array}{*{20}{c}}
			{{D_{xx}}}&{{D_{xy}}}\\
			{{D_{xy}}}&{{D_{yy}}}
	\end{array}} \right],
\end{equation}
where ${D_{xx}}$ and ${D_{yy}}$ denote the second-order orthogonal directional derivatives of ${D(x,y)}$, and ${D_{xy}}$ denotes the second-order cross derivative of ${D(x,y)}$. For each candidate interest point, if the ratio of the two eigenvalues (${\lambda _{\max }}$ and ${\lambda _{\min }}$) is larger than ${\varsigma}$ (e.g., ${\varsigma}$=6), i.e.,

\begin{equation}
	\frac{{{\lambda_{\max }}}}{{{\lambda _{\min }}}} > \varsigma,
\end{equation}
the candidate interest point may be marked as an edge point. Inspired by SIFT~\cite{lowe2004distinctive}, a learned invariant feature transform (LIFT) method~\cite{yi2016lift} was proposed which trains a convolutional neural network on image patches corresponding to the same feature but captured under different ambient conditions. LF-Net~\cite{ono2018lff} was presented which utilizes a fully convolutional network to extract scale-invariant interest points from images. In~\cite{ono2018lff}, a deep architecture and a training strategy were designed for learning local features from scratch based on collections of images without the need of human supervision. In addition to using image pyramids~\cite{lowe2004distinctive}, SEKD~\cite{song2020sekd} combined multi-layer CNN features with different image resolutions, and utilized a mutually reinforcing optimization of interest point detection and obtaining descriptors for improving the performance of interest point detection and its corresponding applications.

In~\cite{10026417}, Zhang et al. investigated and summarized the properties of the second-order isotropic Gaussian directional derivative (SOIGDD) representations of anisotropic-type and isotropic-type blobs and presented a blob detection method. Although tne SOIGDD method achieves good performance on image blob detection, it cannot accurately describe the shape of blobs. The reason is that the second-order isotropic Gaussian directional derivative filters do have not the capability to depict the difference between the anisotropic-type and isotropic-type blobs. To address the aforementioned problem, in this work, the second-order anisotropic Gaussian directional derivative (SOAGDD) filters with multiple scales are utilized for obtaining second-order directional derivatives with multiple scales from input images. Then a novel interest point detection method is proposed which not only has the capability to localize blobs but also has the capability to describe the shape of blobs.

\section{A New Image Blob Detection Using SOAGDD Filters}
In this section, second-order anisotropic Gaussian directional derivative (SOAGDD) filters with multiple scales are introduced. Then a novel interest point detection method based on the SOAGDD filters is proposed.

\subsection{SOAGDD filters}
\begin{figure}[htbp]
	\centering
	\includegraphics[width=1\linewidth]{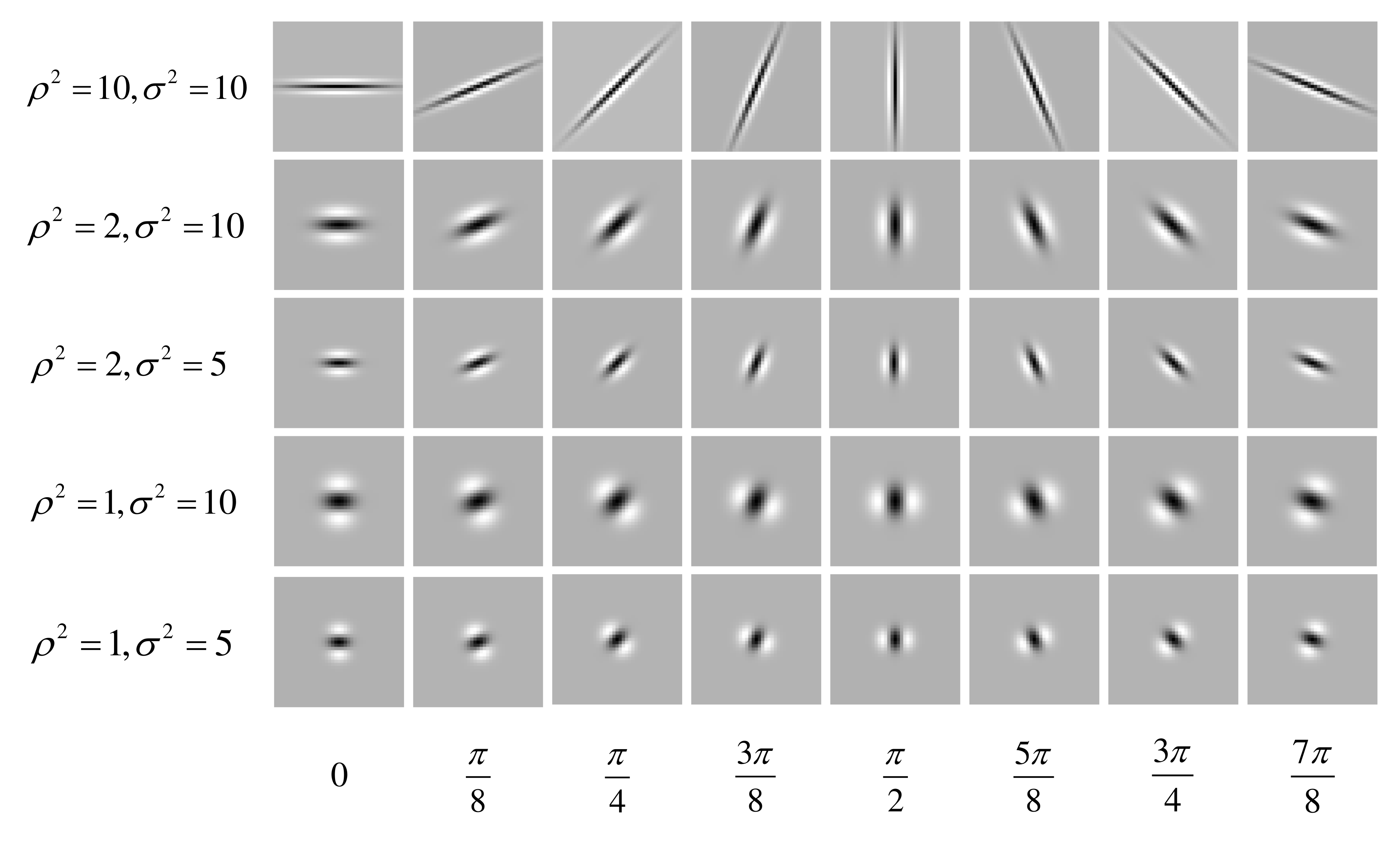}
	\caption{{Examples of SOAGDD filters along eight orientations (${\theta=\frac{{k\pi}}{8}, k = 0,1,\cdots,7}$) with different anisotropic factor $\rho$ and scale factor $\sigma$ .}}
	\label{8}
\end{figure}

In the spatial domain, the anisotropic Gaussian kernel ${{g_{\sigma,\rho,\theta }}\left({x,y} \right)}$ can be represented as~\cite{Shui2012,Zhang2017Noise,shui2012noise,8883063}
\begin{equation}
	\begin{aligned}
		&{g_{\sigma ,\rho ,\theta }}\left( {x,y} \right) =\frac{1}{{2\pi {\sigma ^2}}}\exp \left( { -\frac{1}{{2{\sigma ^2}}}\left[ {x,y} \right]{R_{-\theta }}{C_\rho}{R_\theta }{{\left[ {x,y} \right]}^\top }} \right),\\
		&{R_\theta } = \left[ {\begin{array}{*{20}{c}}
				{\cos \theta }&{\sin \theta }\\
				{ - \sin \theta }&{\cos \theta}
		\end{array}} \right],~{C_\rho } = \left[ {\begin{array}{*{20}{c}}
				{{\rho ^2}}&0\\
				0&{ - {\rho ^2}}
		\end{array}} \right],		
	\end{aligned}
    \label{1}
\end{equation}
where $\top$ denotes matrix transpose, ${(x,y)}$ is a point location in the Cartesian coordinate system, ${\sigma}$ is the scale factor, ${\rho}$ is the anisotropic factor (${\rho}>1$), and $R_{\theta}$ is the rotation matrix with angle rotation ${\theta}$. From Equation~(\ref{1}), the first-order anisotropic Gaussian directional derivative (FOAGDD) filter ${\phi _{\sigma ,\rho ,\theta }}\left( {x,y} \right)$ and the second-order anisotropic Gaussian directional derivative (SOAGDD) filter ${\psi _{\sigma ,\rho ,\theta }}\left( {x,y} \right)$ along the x-axis and then rotating them counterclockwise by ${\theta}$ are derived

\begin{equation}
	\begin{aligned}
		{\phi _{\sigma ,\rho ,\theta }}\left( {x,y} \right)& = \frac{{\partial {g_{\sigma ,\rho }}}}{{\partial \theta }}\left( {{R_\theta }{{\left[ {x,y} \right]}^ \top }} \right)\\
		&=- \frac{{{\rho ^2}\left( {x\cos \theta  + y\sin \theta } \right)}}{{{\sigma ^2}}}{g_{\sigma ,\rho ,\theta }}\left( {x,y} \right),
	\end{aligned}
    \label{2}
\end{equation}
\begin{equation}
	\begin{aligned}
		{\psi _{\sigma ,\rho ,\theta }}\left( {x,y} \right)& = \frac{{\partial {\phi _{\sigma ,\rho }}}}{{\partial \theta }}\left( {{R_\theta }{{\left[ {x,y} \right]}^ \top }} \right) \\
		&= \frac{{{\rho ^2}}}{{{\sigma ^2}}}\left( {\frac{{{\rho ^2}}}{{{\sigma ^2}}}{{\left( {x\cos \theta  + y\sin \theta } \right)}^2} - 1} \right){g_{\sigma ,\rho ,\theta }}\left( {x,y} \right).
	\end{aligned}
    \label{3}
\end{equation}
Examples of SOAGDD filters along eight orientations with different anisotropic factors and scales are shown in Fig.~\ref{8}.
From Equations (\ref{2}) and (\ref{3}), the SOAGDD ${{\psi_{\sigma,\rho,\theta}}\left({x,y}\right)}$ of input image ${I\left({x,y}\right)}$ can be obtained as follows
\begin{equation}
	\label{4}
	\begin{aligned}
	{L_{\sigma,\rho,\theta }}\left( {x,y}\right)& ={\psi_{\sigma,\rho,\theta }}\left({x,y}\right)*I\left({x,y}\right),
	\end{aligned}
\end{equation}
where ${*}$ denotes the convolution operation. It was indicated in~\cite{8883063} that SOAGDD filters have the capability to accurately obtain second-order intensity variation information from images.

\subsection{A new method for image blob detection}
With the advantages of the SOAGDD filters, a new image blob detection method is proposed as follows:

\begin{enumerate}
	\item Image pyramid technique~\cite{lowe2004distinctive} is applied to an input image. The number of layers $n$ of each pyramid is determined as
	\begin{equation}
		\begin{aligned}
		n={\log_2}\left\{{\min \left({M,N}\right)}\right\}-t,~t\in\left[{0,{{\log }_2}\left\{{\min\left({M,N}\right)}\right\}}\right],
		\end{aligned}
		\label{6}
	\end{equation}
where $M$, $N$ are the row and column numbers of the input image, and $t$ represents the size of the minimum image resolution (here $t=2$). Take an house image as an example, its corresponding pyramid images is shown in Fig.~\ref{fig2}.
\begin{figure}[htbp]
	\centering
	\includegraphics[width=0.6\linewidth]{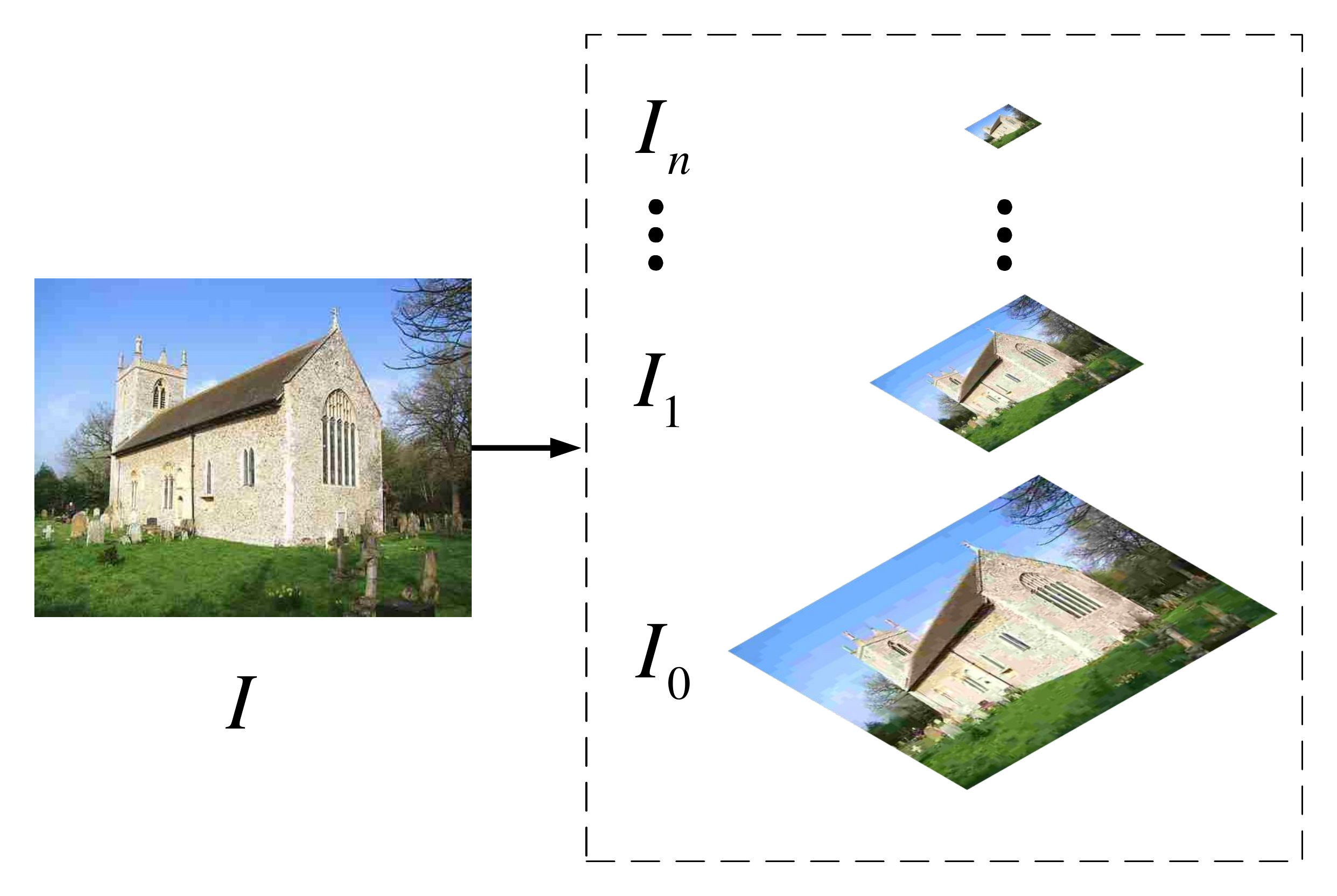}
	\caption{{Example of pyramid images.}}
	\label{fig2}
\end{figure}

%

\item For each image pixel ${({n_x},{n_y})}$, given a scale ${{\sigma_s}~(s=1,\cdots,15)}$ and an anisotropic factor ${{\rho_a}~(a=1,\cdots,5)}$, the product of its corresponding second-order anisotropic directional derivative and the square of scale is accumulated along multiple filtering directions, and then the absolute value of the accumulated result are derived
\begin{equation}
	\begin{aligned}	{\eta_{{\sigma_s}}}\left({{n_x},{n_y}}\right)=\left|{\sum\limits_{k=1}^K{\sigma_s^2{L_{{\sigma_s},{\rho_a},k}}\left({{n_x},{n_y}}\right)}}\right|,
	\end{aligned}
	\label{7}
\end{equation}
where ${K}$ is the number of directions. Pixel ${({n_x},{n_y})}$ will be marked as the center pixel of a candidate blob with scale ${{\sigma_s}}$ if it satisfies:

\begin{equation}
	\begin{aligned}
		{\eta_{{\sigma_s}}}\left({{n_x},{n_y}}\right)&=\max\left\{{\text{set}\left\{{{\eta_{{\sigma_s},{\rho_a}}}\left({{n_x}+u,{n_y}+v}\right)} \right\}} \right\},\\
		{\eta_{{\sigma_s}}}\left({{n_x},{n_y}} \right)&>\max\left\{{\text{set}\left\{{{\eta_{{\sigma_{s+1},{\rho_a}}}}\left({{n_x}+u,{n_y} + v} \right)} \right\}} \right\},\\
		{\eta_{{\sigma_s}}}\left({{n_x},{n_y}}\right)&<\max\left\{{\text{set}\left\{{{\eta_{{\sigma_{s-1},{\rho_a}}}}\left({{n_x}+u,{n_y} +v}\right)}\right\}}\right\}.
	\end{aligned}
	\label{13}
\end{equation}
With ${s=1}$, pixel ${({n_x},{n_y})}$ will be marked as the center pixel of a candidate blob with scale ${\sigma_1}$ only if it satisfies
\begin{equation}
	 \begin{aligned}		{\eta_{{\sigma_s}}}\left({{n_x},{n_y}}\right)&=\max\left\{{\text{set}\left\{{{\eta_{{\sigma_s},{\rho_a}}}\left({{n_x}+u,{n_y}+v} \right)}\right\}}\right\},\\
	{\eta_{{\sigma_s}}}\left({{n_x},{n_y}}\right)& >\max\left\{{\text{set}\left\{{{\eta_{{\sigma_{s+1},{\rho_a}}}}\left({{n_x}+u,{n_y}+v}\right)}\right\}}\right\}.
	\end{aligned}
	\label{9}
\end{equation}
With ${s=15}$, pixel ${({n_x},{n_y})}$ will be marked as the center pixel of a candidate blob with scale ${\sigma_15}$  only if it satisfies with
\begin{equation}
	\begin{aligned}
			{\eta_{{\sigma_s}}}\left({{n_x},{n_y}}\right)&=\max\left\{{\text{set}\left\{{{\eta_{{\sigma_s},{\rho_a}}}\left({{n_x}+u,{n_y}+v}\right)} \right\}}\right\},\\
		{\eta_{{\sigma_s}}}\left({{n_x},{n_y}}\right)&>\max\left\{{\text{set}\left\{{{\eta_{{\sigma_{s-1},{\rho_a}}}}\left({{n_x}+u,{n_y}+v} \right)}\right\}} \right\},
		\end{aligned}
	\label{10}
	\end{equation}
where $\left({{n_x}+u,{n_y}+v}\right)$ represents the neighboring pixels of pixel ${({n_x},{n_y})}$ with ${u\in\left\{{0,\pm 1,\pm 2, \pm 3}\right\}}$ and ${v\in\left\{{0, \pm 1, \pm 2, \pm 3}\right\}}$. Candidate blob is marked as a blob if the blob measure is larger than a given threshold ${{T_b}}$. As shown in Fig. \ref{fig4}, each pixel is compared with all its adjacent pixels to check whether the measure is larger or smaller than that of its adjacent pixels in the image domain and scale domain. The pixel in the middle is compared with 26 other pixels to ensure that extreme pixel is detected in scale space.

\begin{figure}[htbp]
	\centering
	\includegraphics[width=0.4\linewidth]{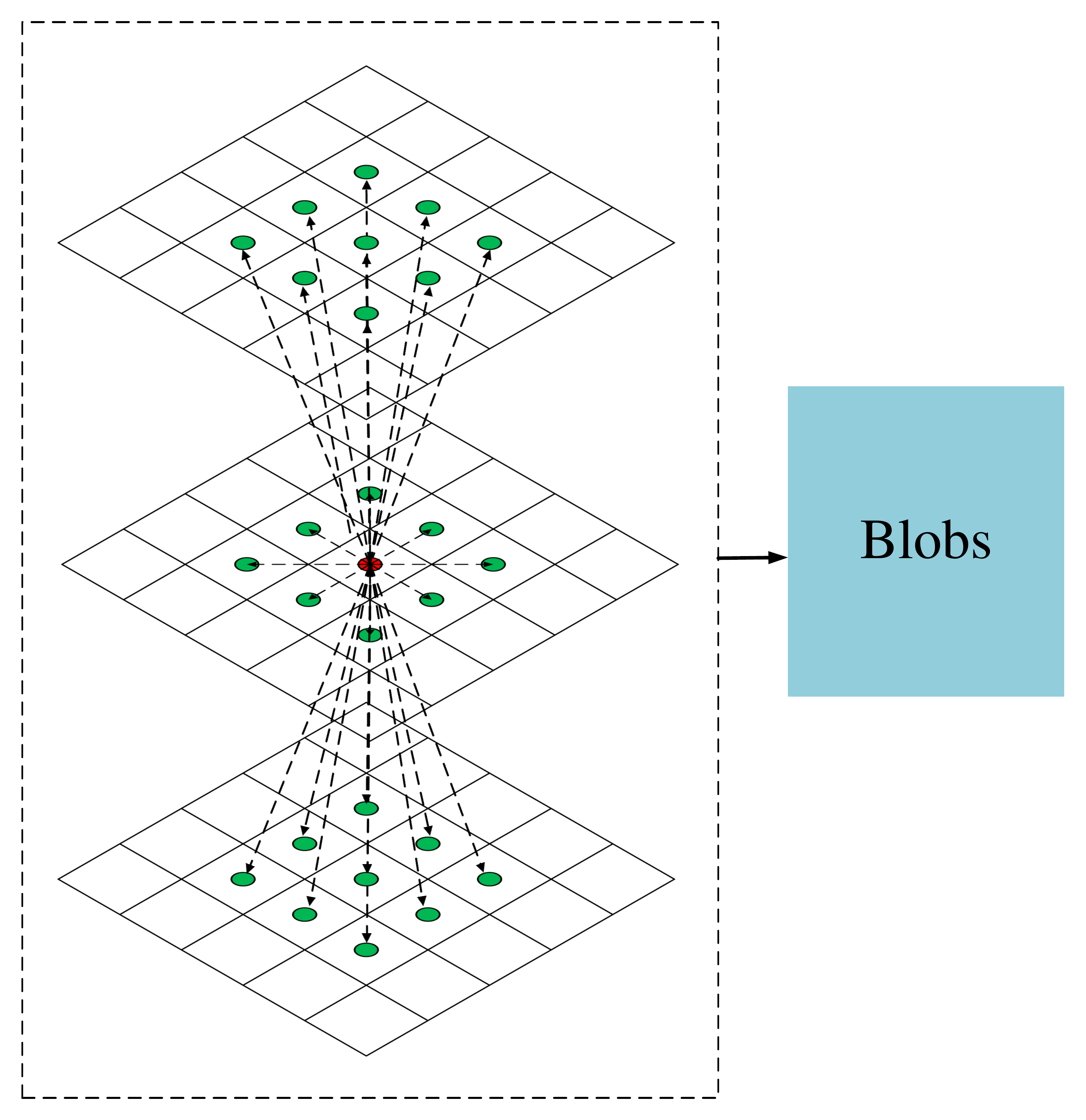}
	\caption{{Illustration of blob selection.}}
	\label{fig4}
\end{figure}
%

\item The direction corresponding to the local maximum SOAGDD along different SOAGDDs is marked as the direction of the short axis. Therefore, the ${{\sigma _s}}$ is marked as the length of the short axis and the ${{\rho _a}}$ is marked as the length of the long axis which are utilized to shape the blob.
\end{enumerate}

\section{Experiments}
\begin{figure}[htp]
	\centering
	\includegraphics[width=1\linewidth]{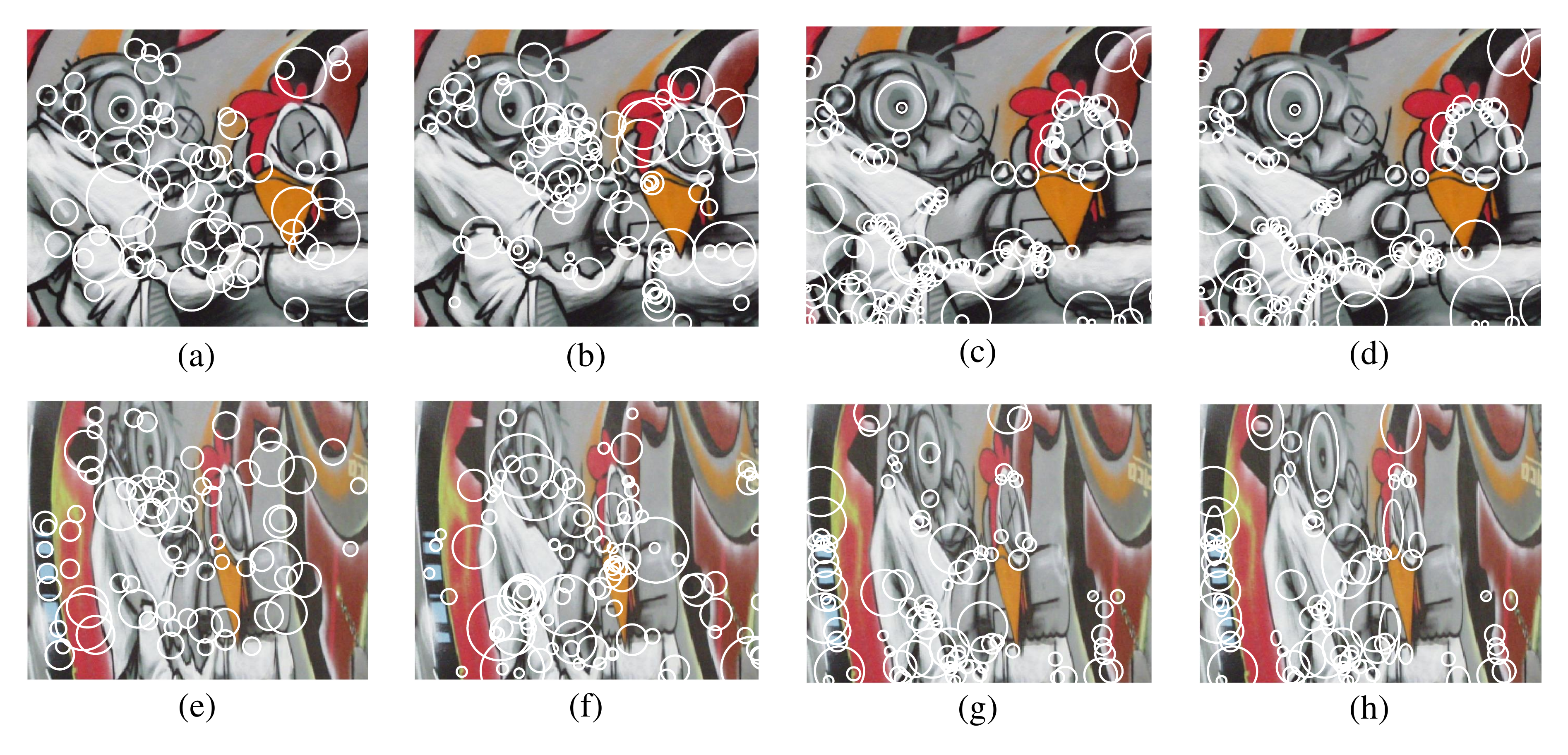}
	\caption{Illustration of different blob detections on the `Graffiti' images. The detection results of SIFT are shown in (a) and (e). The detection results of KAZE are shown in (b) and (f). The detection results of SOIGDD are shown in (ac) and (g). The detection results of proposed method are shown in (d) and (h).}
	\label{tx}
\end{figure}

In this section, the proposed blob detector is compared with three benchmark blob methods (SIFT~\cite{lowe2004distinctive}, KAZE~\cite{alcantarilla2012kaze}, and SOIGDD~\cite{10026417}) by using two images. The parameter settings for the proposed blob detection method are: ${\sigma_s^2\in\left\{{2,\cdots,16}\right\}}$, ${\rho_a^2\in\left\{{1,\cdots,5}\right\}}$, ${K=8}$, and ${{T_b}=223}$. The detection results of the four methods are illustrated in Fig.~\ref{tx}. It can be observed from Fig.~\ref{tx}(d) and (h) that the proposed SOAGDD blob detector has the capability to localize blobs and describe the shape of blobs. This is impossible for the three other state-of-the-art methods~\cite{lowe2004distinctive,alcantarilla2012kaze,10026417}.

\begin{figure}[htbp]
	\centering
	\includegraphics[width=1\linewidth]{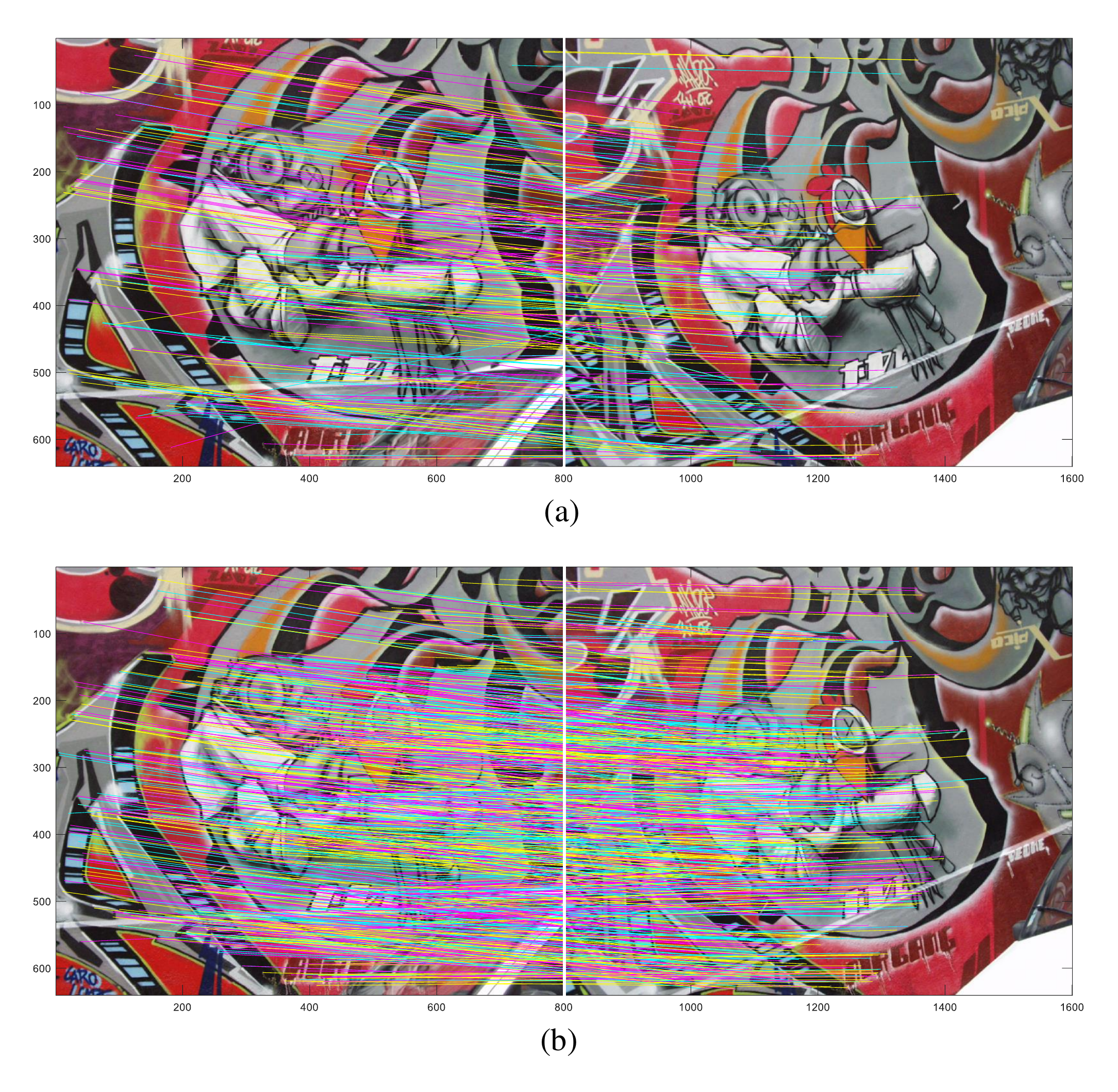}
	\caption{{Image matching. (a) SIFT; (b) The proposed method.}}
	\label{fig5}
\end{figure}

Furthermore, based on SIFT descriptor~\cite{lowe2004distinctive}, the proposed SOAGDD method is compared with the SIFT method on the application of image matching using `Graffiti' images. The SIFT method and the proposed method detect about 1,000 blobs from the images by tuning the threshold respectively. The SIFT method has 323 blob pairs and the proposed method has 659 blob pairs. And their corresponding image matching results are shown in Fig.~\ref{fig5}. It can be found that the proposed method achieve better performance on image matching.

\section{Conclusion}
In this work, the SOAGDD filters with multiple scales are utilized to smooth the input image and obtain second-order anisotropic Gaussian directional derivatives with multiple scales. Based on the SOAGDDs with multiple scales, a new blob detection method is proposed. The proposed method has the capability to localize blobs and describe the shape of blobs. Extensive experiments demonstrate the superiority of our proposed method over state-of-the-art benchmarks based on the localization of interest point detection, shape description of detected interest points, and image matching.

%

\bibliographystyle{llncs}
\bibliography{cite}
\end{document}